%% file: 00_acl_latex.tex
\pdfoutput=1
\documentclass[11pt]{article}

\usepackage[preprint]{acl}

\usepackage{times}
\usepackage{latexsym}
\usepackage{float}
\usepackage{tabularx} 
\usepackage[T1]{fontenc}
\usepackage[utf8]{inputenc}

\usepackage{microtype}

\usepackage{inconsolata}

\usepackage{booktabs}
\usepackage{graphicx}
\usepackage{longtable}

\usepackage{stackengine}
\usepackage{multirow}
\usepackage{graphicx}
\usepackage{wrapfig}
\usepackage{hyperref}
\usepackage{makecell}
\usepackage{booktabs} % For better horizontal lines
\usepackage{array}    % For specifying widths
\usepackage{geometry} % Optional: to adjust page margins if needed

\usepackage{url}            % simple URL typesetting
\usepackage{amsfonts}       % blackboard math symbols
\usepackage{nicefrac} 
\usepackage[figure]{hypcap}
\usepackage{microtype}      % microtypography
\usepackage{xcolor,colortbl}         % colors
\usepackage{times}
\usepackage{latexsym}
\usepackage{multicol}
\usepackage{blindtext}
\usepackage{tabu}
\usepackage{amsmath, bm}

\usepackage{subcaption}
\usepackage{listings} 
\usepackage[most]{tcolorbox}   
\tcbuselibrary{breakable,skins,listings} 
\usepackage{caption}
\usepackage[normalem]{ulem}
\usepackage{soul}
\usepackage[shortlabels]{enumitem}
\usepackage{array}
\usepackage{pgffor}
\usepackage{comment}
\usepackage{textcomp}
\usepackage{amssymb}
\usepackage{pifont}
\usepackage{enumitem}
\usepackage{booktabs}
\usepackage{tabularx}
\usepackage{algorithm}
\usepackage{booktabs}
\usepackage{multirow}
\usepackage{array}
\usepackage{booktabs}
\usepackage{multirow}
\newlist{compactdesc}{description}{1}
\setlist[compactdesc]{%
  leftmargin=1.8em,        % indent for labels
  itemsep=0.4ex,           % space between items
  parsep=0pt,              % additional paragraph separation
  topsep=0.6ex,            % space above the list
  labelsep=0.6em,          % space between label and text
  font=\normalfont\bfseries
}
\usepackage{array}
\usepackage{tabularx}
\newcolumntype{C}{>{\centering\arraybackslash}X}
\usepackage{algorithmic}

\definecolor{lightgreen}{rgb}{0.8, 0.95, 0.8}
\definecolor{lightred}{rgb}{0.95, 0.8, 0.8}
\definecolor{naplesyellow}{rgb}{0.98, 0.85, 0.37}
\definecolor{pastelyellow}{rgb}{0.99, 0.99, 0.59}

\title{Lost in Speech: Benchmarking, Evaluation, and Parsing of Spoken Bilingual Conversational Language Beyond Standard UD Assumptions}

\author{
  Nemika Tyagi$^{1}$\thanks{Equal contribution.} \thanks{Corresponding Author.}\quad
  Olga Kellert$^{1}$\footnotemark[1]\quad
  Holly Hendrix$^{1}$\footnotemark[1]\quad
  Nelvin Licona-Guevara$^1$\footnotemark[1] \quad \\
  \textbf{Justin Mackie}$^1$\quad 
  \textbf{Phanos Kareen}$^1$\quad
  \textbf{Megan Michelle Smith}$^1$ \quad  \\
  \textbf{Tatiana Gallego Hernandez}$^1$ \quad
  \textbf{Samhitha Harish}$^1$ \quad
  \textbf{Chitta Baral}$^1$ \quad \\\\
  $^1$Arizona State University \quad \\
  \small{\texttt{\{ntyagi8, olga.kellert, hbhendri, nliconag\}@asu.edu}}
}
\begin{document}
\maketitle
%\begin{abstract}
%Spoken code-switching (CSW) challenges syntactic parsing in ways not observed in written text. Disfluencies, repetition, ellipsis, and discourse-driven structure routinely violate standard Universal Dependencies (UD) assumptions, causing parsers and large language models (LLMs) to fail despite strong performance on written data. These failures are compounded by rigid evaluation metrics that conflate genuine structural errors with acceptable variation. In this work, we present a systems-oriented approach to spoken CSW parsing. We introduce a linguistically grounded taxonomy of spoken CSW phenomena and \textbf{SpokeBench}, an expert-annotated gold benchmark designed to test spoken-language structure beyond standard UD assumptions. We further propose \textbf{Flex-UD}, an ambiguity-aware evaluation metric, which reveals that existing parsing techniques perform poorly on spoken CSW by penalizing linguistically plausible analyses as errors. We then propose \textbf{DECAP}, a decoupled agentic parsing framework that isolates spoken-phenomena handling from core syntactic analysis. Experiments show that \textbf{DECAP} produces more robust and interpretable parses without retraining and achieves up to \textbf{52.6\%} improvements over existing parsing techniques. Flex-UD evaluations further reveal qualitative improvements that are masked by standard metrics\footnote{Data and source code are available at \url{https://anonymous.4open.science/r/sbench-0C0C}}.
%\end{abstract}

\begin{abstract}
%Spoken code-switching (CSW) presents substantial challenges for syntactic parsing. Conversational bilingual speech often contains disfluencies and discourse-driven structures that complicate dependency parsing under standard Universal Dependencies (UD) assumptions and evaluation practices. In this work, we introduce \textbf{SpokeBench}, an expert-annotated gold benchmark for structurally complex conversational bilingual speech, and a linguistically grounded taxonomy of spoken CSW phenomena. %Moreover, we present a human annotation survey that measures the perceived cognitive and structural difficulty of the phenomena in the taxonomy. 
%We further propose \textbf{Flex-UD}, an ambiguity-aware evaluation metric that distinguishes catastrophic structural errors from acceptable variation, and \textbf{DECAP}, a decoupled agentic parsing framework that separates spoken-phenomena handling from core syntactic analysis. Experiments show that \textbf{DECAP} produces more robust and interpretable parses without retraining and achieves up to \textbf{52.6\%} improvements over existing parsing techniques. Flex-UD evaluations further reveal improvements that remain partially hidden under standard attachment-based metrics\footnote{Data and source code are available at \url{https://anonymous.4open.science/r/sbench-0C0C}}.

Spoken bilingual conversations pose substantial challenges for syntactic parsing because they often include disfluencies and discourse-driven structures that complicate dependency parsing under standard Universal Dependencies (UD) assumptions and evaluation practices. To systematically study these challenges, in this work, we first introduce a linguistically grounded taxonomy of conversational bilingual phenomena, together with \textbf{SpokeBench}, an expert-annotated English-Spanish benchmark for structurally complex speech. To address the limitations of existing evaluation practices, we propose \textbf{Flex-UD}, an ambiguity-aware evaluation metric that distinguishes catastrophic structural failures from linguistically acceptable variations. Finally, we introduce \textbf{DECAP}, a decoupled agentic parsing framework that separates spoken-phenomena handling from core syntactic analysis, enabling robust and interpretable dependency parsing without retraining. Experiments across both proprietary and open-weight LLMs show that \textbf{DECAP} substantially improves performance on complex conversational phenomena and achieves over \textbf{60\%} improvements in UPOS-F1 Score over baselines, while Flex-UD evaluations reveal gains that otherwise remain partially hidden under standard attachment-based metrics\footnote{Data and source code are available at \url{https://github.com/N3mika/LostInSpeech}}.

\end{abstract}

\input{01_introduction}

\input{02_related_works}

\input{03_spokencsw}

\input{04_methodology}

\input{05_experiments}

\input{06_disc_analysis}

\input{07_conclusion}

\bibliography{custom}

\clearpage

\appendix
\input{08_appendix}

\end{document}

%% file: 01_introduction.tex
\section{Introduction}
\label{sec:intro}

\begin{figure}[t!]
    \centering
    \includegraphics[width=7.5cm]{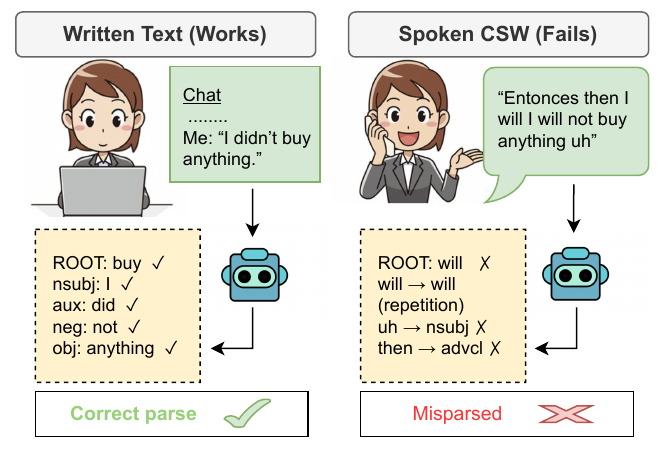}
   \caption{Illustration of the modality gap motivating this work. Parsers and LLMs typically produce well-formed dependency analyses for written text (left), but the same systems often misparse spoken bilingual conversational utterances (right) due to disfluencies, repetition, and discourse phenomena that violate written-text assumptions.}
    \label{fig:fig1}
\end{figure}

Unlike written text, conversational speech frequently contains disfluencies and discourse-driven structures that violate many of the assumptions underlying canonical annotation frameworks such as Universal Dependencies (UD). These phenomena are particularly common in bilingual interaction, where speakers dynamically negotiate structure across languages during code-switching. As illustrated in Figure~\ref{fig:fig1}, models that perform well on written text often fail on conversational bilingual transcripts, producing brittle or linguistically implausible analyses when faced with spoken-language variability and structural ambiguity.

Recent work has shown that large language models (LLMs) can serve as effective tools for syntactic annotation and parsing in multilingual and low-resource contexts, particularly for written text \citep{tian2024large, kellert2025parsing, james2022language, huzaifah2024evaluating}. However, their performance remains understudied and inconsistent for conversational bilingual transcripts, particularly in spoken-language contexts, which introduce ambiguity that is often poorly captured by standard parsing evaluation practices. This degradation cannot be explained by bilingualism or data scarcity alone. Rather, it reflects a mismatch between written-language annotation conventions, rigid evaluation practices, and the incremental, interactional nature of conversational speech \citep{kahane2021annotation, dobrovoljc2022spoken}. As a result, structurally plausible analyses are often penalized as outright parsing failures.
%Constructing gold syntactic annotations for conversational speech is substantially more time-consuming and cognitively demanding than annotation for canonical written text. 
Many parsing failures arise from limited spoken training data and structural assumptions, including non-canonical roots and discourse-level attachments, embedded within existing annotation schemes and evaluation metrics. This suggests that improvements in model scale alone are unlikely to provide a robust or language-agnostic solution. 

%In this work, we argue that progress on spoken code-switched parsing requires rethinking both parsing architectures and evaluation practices, rather than scaling model capacity or data alone. We adopt a systems-oriented perspective that explicitly separates spoken-language phenomena from core syntactic analysis, and we ground this design in an empirical study of the structures that most frequently undermine canonical parsing assumptions. Building on this analysis, we introduce new benchmarks, metrics, and parsing abstractions tailored to the realities of conversational speech. Our contributions are:

In this work, we argue that progress in parsing spoken bilingual conversational language requires jointly rethinking annotation, evaluation, and parsing abstractions rather than treating conversational phenomena as noise around written-language syntax. We adopt a systems-oriented perspective that explicitly separates spoken-language phenomena from core syntactic analysis, grounding this design in an empirical study of the structures that consistently undermine canonical parsing assumptions. Building on this analysis, we introduce new resources, evaluation strategies, and parsing abstractions designed specifically for structurally variable conversational English-Spanish speech. Our contributions are:
\begin{itemize}[topsep=2pt, itemsep=1pt, parsep=0pt, partopsep=0pt]
    \item We design a linguistically grounded \textbf{taxonomy} of spoken bilingual conversational phenomena, distinguishing spoken-specific structures from broader non-canonical constructions and analyzing their structural and cognitive complexity through a \textbf{human annotation survey} measuring annotation difficulty, ambiguity, and cognitive load. These human judgments provide an empirical basis for evaluating and interpreting automatic parsing performance.
    \item We present \textbf{SpokeBench}, an expert-annotated and adjudicated gold benchmark for evaluating dependency parsing on structurally complex bilingual conversational transcripts.
    \item We introduce \textbf{DECAP}, a decoupled agentic parsing framework that improves robustness and interpretability by isolating spoken-language handling from core syntax.
    \item We propose \textbf{Flex-UD}, a weighted, ambiguity-aware evaluation metric that penalizes structurally catastrophic errors more than acceptable variation.

\end{itemize}

Together, these contributions provide a foundation for studying and evaluating syntactic parsing beyond written-language assumptions, highlighting the importance of ambiguity-aware evaluation and speech-informed parsing abstractions for conversational bilingual language.

%% file: 02_related_works.tex
\section{Related Work}
\label{sec:related}

\paragraph{UD Parsing for Non-Canonical and Code-Switched Text.}
UD is a widely adopted framework for cross-linguistic syntactic 
annotation \citep{deMarneffe2021UD, nivre2016ud}, but its application to non-canonical data such as conversational 
speech and intra-sentential bilingual code-switching remains problematic \citep{kellert2025parsing}. Spoken and 
code-switched utterances frequently violate core UD assumptions, including 
clause completeness and canonical head selection, leading to substantial 
annotation ambiguity \citep{ccetinouglu2019challenges, kellert2025parsing, bhat2018universal}. To address these challenges, researchers have developed 
dedicated bilingual treebanks \citep{ccetinouglu2023two, kyle2022dependency, 
pulido2025speak, sanguinetti2023treebanking} and explored alternative 
dependency formalisms such as Surface-Syntactic Universal Dependencies (SUD), 
which modify head assignment to better reflect spoken structure 
\citep{kahane2021annotation, gerdes2019improving}. 
Despite these efforts, prior work has largely focused on documenting annotation 
difficulties rather than enabling robust, scalable parsing or evaluation of 
spoken bilingual data. As a result, current models remain limited in their ability to support accurate syntactic analysis, large-scale computational processing, and systematic evaluation of spoken bilingual language. %Consequently, off-the-shelf parsers trained under standard UD conventions continue to struggle with disfluencies, fragments, and mixed-language constructions, underscoring the need for spoken-aware abstractions beyond existing guidelines.

\begin{table*}[t]
\centering
\scriptsize
\setlength{\tabcolsep}{4pt}
\renewcommand{\arraystretch}{1.1}

\begin{tabular}{p{2.2 cm} p{10.1cm} p{2.7cm}}
\toprule

\textbf{Category} & \textbf{Category Definition} & \textbf{Example} \\

\midrule

\multicolumn{3}{c}{\textbf{Phenomena Shared Across Written and Spoken Conversation}} \\

\midrule

\textbf{Contractions}
& Tokens encoding multiple syntactic units, such as English contractions or Spanish merged forms, which require explicit segmentation during dependency annotation.
& we're \textit{[we are]}, al \textit{[to the]} \\

\textbf{Compound / MWE}
& Multi-word expressions and conventionalized compounds functioning as a single syntactic unit that may obscure internal dependency relations when treated compositionally.
& F\_bat, spring\_break \\

\textbf{Slang / Curses}
& Pragmatically dependent expressions whose syntactic interpretation varies substantially across discourse and cultural context, often resulting in inconsistent annotation and parser behavior.
& shit, papi \textit{[daddy]} \\

\textbf{Enclisis}
& Spanish verbal constructions containing attached clitic pronouns that require morpheme-level decomposition for accurate dependency assignment.
& fíja\textit{te} \textit{[(you) look]}, convirtiéndo\textit{se} \textit{[making itself]} \\

\midrule

\multicolumn{3}{c}{\textbf{Spoken-Specific Conversational Phenomena}} \\

\midrule

\textbf{Repetition}
& Recurrence of words or phrases during incremental speech production, often used to maintain the conversational floor while planning upcoming material. We observe (a.) single-token repetition, (b.) token-separated repetition, and (c.) code-switched repetition across languages.
& (a.) es es \textit{[it's it's]}, (b.) casa uh casa \textit{[house uh house]}; (c.) short cortos \textit{[short short]} %(type c is denoted as repetition+)
\\

\textbf{Discourse Elements}
& Pragmatic markers, interjections, and conversational connectives whose syntactic integration varies across contexts. We distinguish between syntactically integrated discourse elements and pragmatically motivated insertions with weak structural attachment.
& I mean yo tengo \textit{[I mean I have]}\\

\textbf{Ellipsis}
& Structurally incomplete utterances arising from interruption, abandonment, or incremental production. Elliptical constructions frequently omit expected syntactic arguments, complicating dependency assignment and recoverability.
& ah before we; yeah ella se \textit{[yeah she]} \\

\textbf{False Start}
& Mid-utterance reformulations in which a speaker abandons an initiated structure and replaces it with a revised continuation, often reflecting online adjustment of bilingual grammatical planning and requiring inference over partially realized syntactic structure and speaker intent.
& pero I lo I use it \textit{[but I it I use it]} \\

\textbf{Filler Words}
& Floor-holding vocalizations carrying minimal semantic content but remaining interactionally important in conversational speech, often with weak syntactic integration.
& uh, um, eh \\

\bottomrule
\end{tabular}

\caption{Taxonomy of conversational bilingual phenomena identified in the Miami Corpus. Phenomena are grouped by whether they occur in both written and spoken discourse or are primarily associated with conversational speech.}
\label{tab:taxonomy_main}
\end{table*}
\paragraph{LLMs for Annotation and Parsing.}

Large Language Models (LLMs) have recently been explored as parsers and annotation tools, showing that they can produce syntactic analyses for written text in zero-shot or few-shot settings \citep{zhang2025self}, and achieve competitive performance when fine-tuned \citep{bai2025constituency}. These capabilities motivate their use in bootstrapping annotations and generating synthetic treebanks, even for low-resource languages. However, consistent limitations remain. Beyond the well-documented challenges in written language, studies indicate that LLM performance degrades significantly with flexible, noisy, or conversational input, including spoken and CSW data \citep{de2024code}. Prior works have explored alternative parsing architectures, such as sequence-labeling approaches \citep{gomez2024dancing, imran2024syntax} or incorporating syntactic information into downstream tasks \citep{imran2025synner}. While these models show promise for certain spoken-language tasks (e.g., detection or translation) \citep{james2022language, huzaifah2024evaluating}, ability to robustly parse fragmented or ambiguous syntax of real speech remains inconsistent.

\paragraph{Evaluation Limitations.}
Standard dependency parsing metrics such as Labeled and Unlabeled Attachment Score (LAS and UAS) rely on strict matching against a single gold tree, making them poorly suited to conversational and structurally ambiguous language where multiple analyses may be linguistically plausible. Prior work has shown that these metrics often conflate genuine parsing failures with acceptable structural variation \citep{sterner2025minimal, sterner2025code}, motivating the need for ambiguity-aware evaluation.

%% file: 03_spokencsw.tex
\section{Spoken Bilingual Conversation: A Non-Canonical Parsing Domain}
\label{sec:spoken}
In this section, we examine spoken bilingual conversational text as a complex parsing domain. First, we develop a linguistically grounded taxonomy of conversational phenomena, followed by a human-perception survey investigating their structural and cognitive complexity. We then introduce \textbf{SpokeBench}, a gold benchmark for evaluating parsing performance under spoken-text conditions.

\subsection{Taxonomy of Spoken Bilingual Conversational Phenomena}
%Conversational bilingual speech exhibits a range of non-canonical phenomena, including disfluencies, repetitions, and structurally incomplete utterances that remain underrepresented in standard text corpora and annotation guidelines. 
To identify the primary sources of parsing difficulty in conversational bilingual language, we conducted a data-driven analysis of English-Spanish utterances from the Miami Corpus \citep{deuchar2014building}, a widely used English-Spanish bilingual conversational dataset. Approximately 2,800 sentences were independently examined by trained linguists, who analyzed and identified conversational phenomena that complicate non-canonical dependency parsing. The consolidation of these observations yielded a linguistically grounded taxonomy, motivated by recurrent parser failures and annotation difficulties.
Many of these phenomena can in principle be represented within existing UD and SUD frameworks, but conversational bilingual speech substantially increases annotation ambiguity and structural variability in practice. Our goal is therefore not to replace existing annotation schemes, but to systematically analyze how these conversational phenomena interact with bilingual speech and affect parser behavior, annotation consistency, and evaluation.

%\paragraph{Category Definitions}
We organize the identified phenomena into two groups: (i) phenomena occurring in both written and spoken discourse but becoming structurally more challenging in bilingual conversational settings, and (ii) phenomena strongly associated with spontaneous conversational speech and incremental language production. The definitions and examples are provided in Table \ref{tab:taxonomy_main}.

\subsection{SpokeBench: A Gold Benchmark for Spoken Conversational Parsing}
\paragraph{Subset Selection Strategy.}
Based on the conversational phenomena identified in the Miami Corpus, we construct \textbf{SpokeBench}, a gold benchmark for evaluating dependency parsers on structurally complex bilingual conversational transcripts. %Existing parsing resources provide limited coverage of conversational spoken-language variability, particularly for bilingual and structurally non-canonical utterances. 
Since annotating the full set of over 2,800 sentences was infeasible due to the cost and ambiguity of conversational speech annotation, we selected a curated subset balancing coverage of spoken phenomena and structural complexity. Most of the sentences have overlapping phenomena and some categories are extremely rare, making it hard to strictly isolate sentences for each category in the taxonomy. Therefore, we constructed SpokeBench with a balance of isolated dominant categories (Repetition, Contraction [EN \& ES], Ellipsis, Discourse, Fillers) and categories (denoted with `+') that consist of a dominant phenomenon along with 1 or 2 rare phenomena, such as compound, slang, or false start. The category ``Complex'' refers to the sentences that have 3+ phenomena and ``Others'' refer to sentences unique in their presentation but do not fit in the above categories.
%Because many utterances exhibit overlapping phenomena, sentences were stratified by both dominant phenomenon and complexity level rather than assigned to a single category. 
The resulting benchmark contains 160 sentences spanning ten categories; its composition is summarized in Appendix~\ref{app:spokebench}.
%Table~\ref{tab:spokebench_subset} 
Unlike most existing UD treebanks, which primarily focus on annotation coverage or multilingual representation, SpokeBench is explicitly designed to evaluate parser robustness under structurally ambiguous conversational conditions.

\paragraph{Annotation Procedure.} The
Annotations in \textbf{SpokeBench} follow a modified version of UD guidelines adapted for spoken bilingual conversational data. Standard UD conventions were applied where possible, with additional rules introduced for repetitions, contractions, discourse markers, fillers, and ellipsis. Each sentence was independently annotated by at least two linguistically trained annotators (seven in total), and disagreements were evaluated using a graded acceptability protocol distinguishing acceptable variation from substantive annotation errors. Borderline or unacceptable cases were resolved through expert adjudication, yielding a consensus gold annotation for each sentence. To evaluate annotation consistency, we calculated Cohen’s Kappa score for a randomized subset of 50 sentences, which resulted in 0.798. Appendix~\ref{app:guidelines} contains the detailed annotation guidelines and review procedures. 

\begin{figure}[t]
    \centering
    \includegraphics[width=\columnwidth]{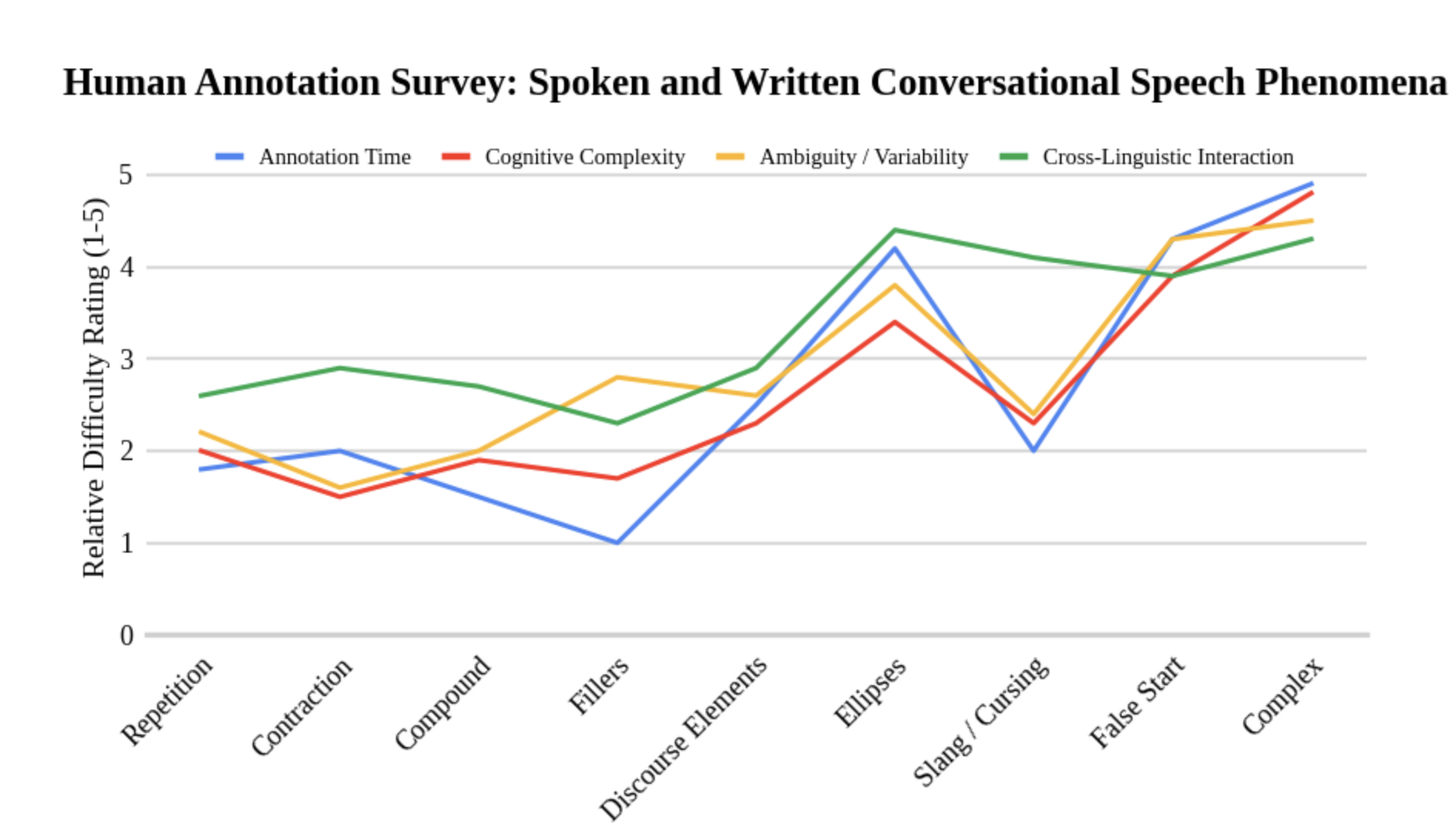}
    \caption{Average Likert-scale ratings from expert annotators across four dimensions: annotation time, cognitive complexity, ambiguity/variability, and cross-linguistic interaction.}
    \label{fig:survey}
\end{figure}

\label{sec:decap}
\begin{figure*}[t]
  \centering
  \includegraphics[width=\textwidth]{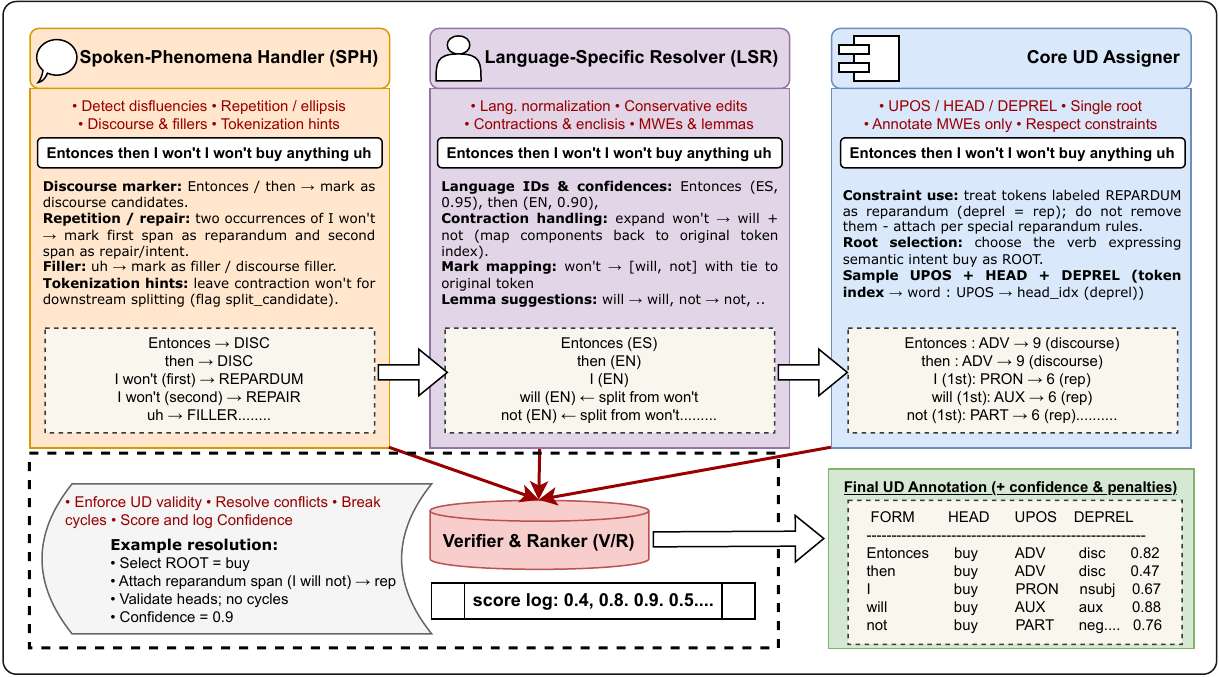}
  \caption{Overview of the \textbf{DECAP} framework for spoken code-switching parsing, illustrated with a running example. The input utterance (“\textit{Entonces then I won’t I won’t buy anything uh}”) is processed. The \textbf{Spoken-Phenomena Handler (SPH)} detects disfluencies (e.g., repetition, discourse markers, fillers) and provides tokenization hints. The \textbf{Language-Specific Resolver (LSR)} applies conservative, language-aware normalization, and contraction splitting. The \textbf{Core UD Assigner} constructs a dependency parse under these constraints, preserving reparanda and enforcing a single root. Finally, the \textbf{Verifier and Ranker (V/R)} enforces global UD validity and outputs a final annotation with confidence and penalty scores, demonstrating incremental resolution of ambiguity.}

  \label{fig:mainfig}
\end{figure*}

\subsection{Human Perception Survey}
Following the annotation process, we conducted a human perception survey involving the expert annotators who participated in the construction and adjudication of \textbf{SpokeBench}. Annotators evaluated each conversational phenomenon on a five-point Likert scale (1 = least difficult, 5 = most difficult) across four dimensions: annotation time, cognitive complexity, ambiguity/variability, and cross-linguistic interaction. As shown in Figure~\ref{fig:survey}, contractions, filler words, and localized repetitions were generally perceived as less difficult to annotate, while ellipsis, discourse elements, and false-start constructions received the highest difficulty ratings across nearly all dimensions due to their reliance on contextual interpretation, structural recoverability, and speaker-intent inference. These findings suggest that conversational phenomena involving greater ambiguity and discourse-level reasoning present substantially greater challenges for human annotators. We hypothesize that the same phenomena will also present systematic challenges for automatic parsers, particularly for dependency attachment and structural recoverability. This hypothesis is tested experimentally in Section 5. %add reference here nem

\begin{comment}
\begin{table}[t]
\centering
\small
\begin{tabular}{l r}
\hline
\textbf{Phenomenon} & \textbf{Frequency} \\
\hline
Repetition & 99 \\
Break of Thought & XX \\
Ellipsis & XX \\
Discourse Elements & XX \\
Filler Words & XX \\
Slang / Curse Words & XX \\
Contractions & XX \\
Enclisis & XX \\
Compound Words & XX \\
\hline
\end{tabular}
\caption{Distribution of spoken code-switching phenomena across the analyzed subset of the Miami Corpus.}
\label{tab:taxonomy_dist}
\end{table}
\end{comment}

%% file: 04_methodology.tex
\section{DECAP: A Decoupled Agentic Framework for Spoken Bilingual Conversation Parsing}
\label{sec:methd}

In this section, we introduce \textbf{DECAP} (DECoupled Agentic Parser), a modular framework for parsing conversational bilingual transcripts. Rather than treating conversational variability within a single monolithic parser, DECAP separates spoken-phenomena handling, language-specific normalization, syntactic structure assignment, and global validation into specialized stages. The framework is guided by three principles: (i) \textit{instruction grounding}, leveraging LLMs’ ability to follow explicit linguistic constraints; (ii) \textit{modularity}, isolating distinct sources of conversational complexity; and (iii) \textit{extensibility}, enabling adaptation across languages, phenomena, and LLM families without retraining. Figure~\ref{fig:mainfig} illustrates the overall architecture.

\paragraph{Spoken-Phenomena Handler (SPH).}
The SPH agent identifies conversational phenomena prior to syntactic parsing, including repetitions, ellipsis, discourse markers, fillers, contractions, and MWEs, and proposes minimal tokenization edits where required. Rather than directly assigning dependency relations, SPH provides lightweight structural guidance and tokenization hints that constrain downstream parsing decisions and reduce error propagation across later stages.

\paragraph{Language-Specific Resolver (LSR).}
This agent applies conservative language-aware normalization to the output of SPH, including contraction expansion, multiword-expression validation, and lemma normalization. The agent is designed to minimize unnecessary structural repairs while preserving conversational structure and English-Spanish language variations. By isolating language-dependent operations, LSR improves robustness across multilingual and code-switched input.

\paragraph{Core UD Structure Assigner.}
The Core UD Assigner produces dependency annotations over the normalized representation generated by previous agents. It assigns UPOS tags, dependency heads, and dependency relations while enforcing standard UD structural constraints such as single-root validity. Conversational phenomena identified by earlier stages are preserved rather than reconstructed into canonical written-language forms. This stage focuses exclusively on syntactic structure, deferring validation and correction to the final agent.

\paragraph{Verifier and Ranker (V/R).}
The V/R agent integrates the outputs of all previous stages and performs global structural validation, resolving issues such as invalid heads, dependency cycles, and conflicting annotations. In addition to producing sheet-ready annotations, V/R assigns per-token confidence and penalty scores, yielding an interpretable record of uncertainty and intervention. This final stage guarantees that DECAP outputs are structurally valid, auditable, and suitable for downstream evaluation.

\paragraph{Interaction protocol.}
DECAP applies a fixed, deterministic sequence of agents to each sentence, with each stage consuming the structured output of the previous one and enforcing increasingly global constraints, as detailed in the Appendix. \ref{app:int}.

%% file: 05_experiments.tex
\section{Experiments}

%\paragraph{Evaluation Metrics}
%System outputs are evaluated using standard UD metrics, including UAS, LAS, CLAS, and UPOS-based attachment scores, as well as the proposed FLEX-UD metric. Standard metrics are computed against the gold annotations directly, while FLEX-UD provides a severity-aware, ambiguity-tolerant evaluation that produces both component scores and a single aggregated score.

\subsection{Experimental Setup}

\paragraph{Dataset.}

All experiments are conducted on \textbf{SpokeBench}, the benchmark introduced in Section~\ref{sec:spoken}. We evaluate systems on a curated subset of 160 sentences selected to balance conversational phenomena and structural complexity. Gold Universal Dependency annotations are used across all experiments to ensure controlled and directly comparable evaluation.

\paragraph{Models.}

We evaluate four parser configurations spanning both traditional and LLM-based approaches. As a traditional baseline, we use the \textbf{Trankit} parser, a transformer-based multilingual dependency parsing toolkit built on XLM-RoBERTa \citep{nguyen2021trankit}. It supports multilingual UD parsing through contextualized transformer representations and automatic language identification, but is trained on monolingual treebanks only. We then evaluate the proposed \textbf{DECAP} framework instantiated with three different LLM families: GPT-4.1 (version 2025-04-14), \texttt{meta-llama/Llama-3.1-8B-Instruct}, and \texttt{meta-llama/Llama-3.1-70B-Instruct}. All models are evaluated in a zero-shot setting with deterministic decoding (temperature = 0), and none are fine-tuned on \textbf{SpokeBench} or Miami Corpus. This setup allows us to evaluate the extent to which the DECAP framework generalizes across both proprietary and open-weight LLM architectures.

\subsection{Evaluation Metrics}

\paragraph{Standard UD Metrics and Limitations.}
We evaluate parser performance using four standard dependency parsing metrics: Unlabeled Attachment Score (UAS), Labeled Attachment Score (LAS), Content-Labeled Attachment Score (CLAS), and UPOS-F1 Score (UP-F1). UAS measures unlabeled head attachment accuracy, LAS additionally requires correct dependency labels, and CLAS evaluates labeled dependency accuracy over core content relations only. We additionally report UPOS-F1 to evaluate morphosyntactic tagging independently from dependency structure assignment. While these metrics are effective for canonical written text, they are less well suited to conversational bilingual speech, where disfluencies, ellipsis, discourse markers, and structurally incomplete utterances often permit multiple linguistically plausible analyses. As a result, strict gold-tree matching may penalize acceptable structural variation alongside genuine parsing errors.

\begin{table*}[t]
\centering
\scriptsize
\setlength{\tabcolsep}{3pt}
\renewcommand{\arraystretch}{1.08}
\resizebox{\textwidth}{!}{%
\begin{tabular}{lcccccccccccccccc}
\toprule

\multirow{2}{*}{\textbf{Category}} 
& \multicolumn{4}{c}{\textbf{TRANKIT Bilingual}} 
& \multicolumn{4}{c}{\textbf{DECAP Llama 3 8B}} 
& \multicolumn{4}{c}{\textbf{DECAP Llama 3 70B}} 
& \multicolumn{4}{c}{\textbf{DECAP GPT-4.1}} \\

\cmidrule(lr){2-5}
\cmidrule(lr){6-9}
\cmidrule(lr){10-13}
\cmidrule(lr){14-17}

& LAS & UAS & CLAS & UP-F1
& LAS & UAS & CLAS & UP-F1
& LAS & UAS & CLAS & UP-F1
& LAS & UAS & CLAS & UP-F1 \\

\midrule
Repetition     
& 5.3 & 7.7 & 3.3 & 12.4
& 9.8 & 15.2 & 7.8 & 24.0
& 25.9 & 48.2 & 18.8 & 65.8
& 42.5 & 64.7 & 26.7 & 84.7 \\
Repetition+    
& 3.9 & 6.1 & 2.1 & 10.3
& 8.4 & 13.3 & 5.9 & 21.4
& 21.3 & 41.6 & 15.6 & 58.3
& 38.9 & 49.3 & 29.3 & 81.6 \\
Contr. (EN)    
& 4.2 & 7.4 & 1.3 & 8.7
& 1.6 & 4.6 & 0.3 & 28.6
& 21.5 & 37.7 & 11.8 & 57.3
& 31.7 & 48.2 & 25.5 & 83.5 \\
Contr. (ES)    
& 3.4 & 8.3 & 1.2 & 9.4
& 0.8 & 3.7 & 0.1 & 18.8
& 28.4 & 46.3 & 20.5 & 63.1
& 42.8 & 51.7 & 33.8 & 82.6 \\
Ellipsis       
& 1.7 & 4.3 & 0.3 & 13.7
& 10.4 & 14.6 & 7.5 & 29.3
& 34.8 & 50.1 & 22.9 & 61.8
& 33.6 & 43.6 & 27.1 & 81.4 \\
Ellipsis+      
& 0.4 & 2.1 & 0.0 & 10.6
& 7.3 & 11.5 & 3.1 & 17.9
& 31.5 & 48.3 & 17.9 & 55.8
& 45.7 & 59.4 & 26.5 & 81.9 \\
Discourse      
& 1.1 & 4.1 & 0.1 & 11.2
& 5.9 & 8.9 & 1.9 & 25.5
& 25.7 & 38.9 & 19.3 & 58.7
& 31.2 & 40.1 & 21.2 & 86.2 \\
Discourse+     
& 0.2 & 1.7 & 0.0 & 7.8
& 6.8 & 7.6 & 1.6 & 20.6
& 31.6 & 40.8 & 21.0 & 51.6
& 35.4 & 53.9 & 30.1 & 78.7 \\
Complex        
& 1.8 & 4.8 & 1.3 & 9.1
& 3.6 & 5.2 & 0.3 & 18.9
& 28.1 & 43.3 & 21.6 & 53.1
& 32.9 & 41.8 & 27.3 & 80.5 \\
Fillers        
& 0.0 & 0.3 & 0.0 & 2.3
& 1.6 & 3.4 & 1.2 & 15.8
& 30.3 & 47.9 & 26.9 & 59.2
& 31.6 & 38.7 & 17.8 & 77.8 \\
Others         
& 7.6 & 10.6 & 6.3 & 8.7
& 8.7 & 14.9 & 15.3 & 16.2
& 32.8 & 41.7 & 24.3 & 58.8
& 20.1 & 31.6 & 15.9 & 62.8 \\
\midrule
\textbf{Overall}
& \textbf{2.7} & \textbf{5.2} & \textbf{1.6} & \textbf{9.4}
& \textbf{6.2} & \textbf{9.7} & \textbf{4.5} & \textbf{20.8}
& \textbf{28.6} & \textbf{43.8} & \textbf{20.1} & \textbf{57.6}
& \textbf{35.0} & \textbf{47.4} & \textbf{25.6} & \textbf{79.2} \\
\bottomrule
\end{tabular}%
}
\caption{Performance of parsers tested across spoken bilingual conversational categories under standard UD metrics. UP-F1 denotes UPOS-F1 Score}
\label{tab:standard_metrics_main}
\end{table*}

\begin{table*}[t]
\centering
\scriptsize
\setlength{\tabcolsep}{3pt}
\renewcommand{\arraystretch}{1.05}
\resizebox{\textwidth}{!}{%
\begin{tabular}{lcccccccccccccccc}
\toprule

\multirow{2}{*}{\textbf{Category}}
& \multicolumn{4}{c}{\textbf{TRANKIT Bilingual}}
& \multicolumn{4}{c}{\textbf{DECAP Llama 3 8B}}
& \multicolumn{4}{c}{\textbf{DECAP Llama 3 70B}}
& \multicolumn{4}{c}{\textbf{DECAP GPT-4.1}} \\

\cmidrule(lr){2-5}
\cmidrule(lr){6-9}
\cmidrule(lr){10-13}
\cmidrule(lr){14-17}

& UPOS & HEAD & DEP & Flex-UD
& UPOS & HEAD & DEP & Flex-UD
& UPOS & HEAD & DEP & Flex-UD
& UPOS & HEAD & DEP & Flex-UD \\

\midrule
Repetition
& 18.9 & 14.2 & 11.3 & 13.4
& 38.1 & 21.4 & 23.8 & 25.6
& 70.8 & 51.3 & 57.6 & 60.2
& 86.9 & 68.2 & 75.6 & 78.3 \\
Repetition+
& 12.3 & 10.3 & 8.9 & 11.6
& 29.4 & 17.4 & 19.3 & 23.8
& 67.2 & 49.4 & 54.1 & 57.1
& 88.5 & 60.1 & 68.9 & 71.4 \\
Contr. (EN)
& 13.8 & 8.3 & 10.7 & 12.3
& 38.6 & 8.3 & 10.3 & 16.4
& 68.1 & 42.7 & 48.9 & 53.8
& 84.2 & 66.1 & 71.2 & 73.7 \\
Contr. (ES)
& 15.6 & 7.1 & 11.4 & 12.7
& 30.2 & 7.9 & 8.5 & 15.1
& 73.4 & 46.5 & 53.7 & 56.4
& 87.1 & 70.9 & 77.1 & 81.2 \\
Ellipsis
& 21.0 & 14.6 & 7.6 & 14.4
& 37.4 & 17.6 & 20.9 & 24.7
& 72.8 & 48.3 & 51.9 & 55.9
& 84.0 & 59.8 & 73.4 & 75.6 \\
Ellipsis+
& 16.2 & 7.8 & 5.4 & 9.2
& 26.8 & 13.4 & 18.3 & 19.5
& 69.2 & 42.1 & 48.2 & 52.7
& 83.6 & 61.6 & 68.1 & 73.2 \\
Discourse
& 16.3 & 12.1 & 8.0 & 11.1
& 34.1 & 17.0 & 17.4 & 21.3
& 64.3 & 50.6 & 53.4 & 54.8
& 89.4 & 57.5 & 70.6 & 73.9 \\
Discourse+
& 11.7 & 8.2 & 6.2 & 8.5
& 30.5 & 11.3 & 14.1 & 17.7
& 62.8 & 43.1 & 52.1 & 54.6
& 81.7 & 58.2 & 68.9 & 70.6 \\
Complex
& 14.9 & 12.9 & 9.3 & 12.3
& 27.6 & 8.2 & 9.3 & 12.8
& 66.3 & 45.7 & 51.0 & 53.1
& 84.1 & 59.8 & 63.7 & 66.2 \\
Fillers
& 4.5 & 2.3 & 3.1 & 3.6
& 8.6 & 0.6 & 2.6 & 3.1
& 72.5 & 34.5 & 53.6 & 57.7
& 82.4 & 53.1 & 72.5 & 75.1 \\
Others
& 10.6 & 13.8 & 13.8 & 15.7
& 24.7 & 21.5 & 23.4 & 23.8
& 70.6 & 43.6 & 55.1 & 58.3
& 78.7 & 57.1 & 70.1 & 72.3 \\
\midrule
\textbf{Overall}
& \textbf{13.8} & \textbf{10.3} & \textbf{8.7} & \textbf{11.4}
& \textbf{29.1} & \textbf{13.5} & \textbf{15.8} & \textbf{18.8}
& \textbf{68.4} & \textbf{45.1} & \textbf{52.5} & \textbf{55.7}
& \textbf{84.2} & \textbf{60.6} & \textbf{70.0} & \textbf{72.8} \\
\bottomrule
\end{tabular}%
}
\caption{Component-wise evaluation of the linguistically central Flex-UD components (UPOS, HEAD, and DEPREL) together with the aggregate Flex-UD score across spoken bilingual categories for four parser settings.}
\label{tab:flexud_main}
\end{table*}

\paragraph{Flex-UD Metric.}
To better account for structural ambiguity in conversational speech, we introduce \textbf{Flex-UD} (Flexible Evaluation for Universal Dependencies), a composite severity-aware evaluation metric designed for spoken bilingual parsing. Rather than relying exclusively on exact tree matching, Flex-UD evaluates parses across multiple complementary dimensions, including accurate tokenization, ID assignments, UPOS assignments, dependency heads, and dependency relations. The metric is designed to: 
(i) tolerate linguistically acceptable variation, 
(ii) distinguish catastrophic structural errors from minor deviations, and 
(iii) provide interpretable diagnostics for qualitative error analysis. Illustrative examples of error severity and penalty assignment are provided in Appendix~\ref{app:flexud_examples}.

\paragraph{Formal Definition.}

The Flex-UD metric evaluates parses across five component scores 
\(S = \{s_{\text{Split}}, s_{\text{ID}}, s_{\text{UPOS}}, s_{\text{HEAD}}, s_{\text{DEPREL}}\}\), 
corresponding respectively to tokenization accuracy, ID assignment, UPOS agreement, dependency head assignment, and dependency relation assignment. Each score is normalized to the range \([1,100]\) and assigned a weight 
\(w = \{0.10, 0.05, 0.15, 0.25, 0.45\}\), 
placing greater emphasis on dependency structure and relation accuracy. The weighted aggregate score is computed as
\[
\mathrm{raw} = \sum_i w_i s_i.
\]
To account for severe structural failures, Flex-UD applies a bounded penalty term \(P \in [0,1]\), where larger values correspond to more severe parsing inconsistencies. Catastrophic errors such as invalid dependency heads, unresolved cycles, severe tokenization mismatches, or structurally incompatible alignments receive substantially larger penalties than minor linguistic variations such as discourse attachment ambiguity or near-equivalent dependency relations.
The final Flex-UD score is therefore computed as:
\[
\mathrm{Flex-UD} = \mathrm{round}\big(\mathrm{raw}(1 - P)\big).
\]
The final goal of creating Flex-UD is not to conflate model performance by using more tolerant annotations, but to capture how structurally and linguistically reasonable the produced parse is overall. In addition to the aggregate score, Flex-UD outputs component-wise explanations, token alignments, and diagnostic notes to support qualitative analysis and model comparison.
%Although Flex-UD is more tolerant of linguistically acceptable variation than strict attachment-based metrics, it is not designed to systematically inflate scores. Instead, the metric aims to distinguish catastrophic structural failures from partial or locally plausible parses that remain informative despite deviations from a single gold dependency tree.

%% file: 06_disc_analysis.tex
\section{Results and Discussion}
\subsection{Overall Parsing Performance}
%\label{sec:results}

The combined results from Tables~\ref{tab:standard_metrics_main} and~\ref{tab:flexud_main} reveal a consistent pattern across all evaluated systems and linguistic categories. Overall, the stronger systems, especially the DECAP GPT-4.1 and DECAP Llama-3-70B models, substantially outperform the weaker systems across all parsing and tagging metrics, indicating major improvements in robustness to non-canonical and discourse-rich language. However, even the best-performing systems continue to struggle with informal conversational phenomena such as fillers, contractions, ellipsis, and discourse-level structures, demonstrating that syntactic analysis of interactional language remains a difficult problem.

Across both tables, lexical and morphosyntactic recognition is consistently more robust than full syntactic parsing. UPOS tagging and Flex-UD achieve substantially higher scores than strict dependency metrics such as LAS and UAS, suggesting that the systems can often recognize word categories and recover partial local structure even when they fail to reconstruct full hierarchical dependency relations. This difference is especially visible in Table~\ref{tab:flexud_main}, where Flex-UD scores remain relatively high despite lower head and dependency attachment scores, indicating latent structural robustness not captured by strict UD evaluation alone. These findings replicate the general tendency observed in \citep{kellert2025parsing}, while extending the analysis to different parser architectures and LLM families.

\subsection{Human vs Parser Difficulty}

The survey results shown in Figure~\ref{fig:survey} strongly support the interpretation that human annotation difficulty and parser difficulty do not always align. Human annotators rated ellipsis and highly complex constructions as the most cognitively demanding categories, with these phenomena receiving the highest scores across annotation time, ambiguity, and cognitive complexity. By contrast, fillers and contractions were rated as comparatively easier and faster to annotate despite their non-canonical spoken character. This pattern differs substantially from the parser evaluation results in Tables~\ref{tab:standard_metrics_main} and~\ref{tab:flexud_main}, where fillers and contractions produced some of the lowest parsing scores across systems, while several elliptical constructions were handled comparatively better by the stronger models.

This discrepancy suggests an important distinction between human linguistic processing and automatic dependency parsing. Human annotators appear to rely heavily on discourse-pragmatic expectations and interactional competence when interpreting fillers and contractions, making these phenomena relatively easy to process despite their reduced or fragmented form. Current parsing systems, however, may lack sufficient exposure to such interactional phenomena during training, since fillers and spoken disfluencies remain relatively underrepresented in standard treebanks and written corpora. Elliptical constructions, by contrast, seem to pose slightly fewer difficulties for parsers, possibly because certain analogous forms of ellipsis or omission are also attested in written language and therefore occur more frequently in training data. This partial overlap may allow models to generalize somewhat more effectively to elliptical structures than to highly interactional spoken phenomena such as fillers and contractions. The results therefore point to a broader cognitive and annotation-level mismatch between human dependency parsing strategies and the statistical learning biases of contemporary NLP systems.

\subsection{Implications for UD Annotation and Evaluation.}

Even though parsers show slightly greater robustness to certain elliptical constructions than to fillers, ellipsis and discourse-level phenomena continue to produce substantial degradation across systems. Our findings expose a tension between surface-oriented approaches such as Surface-Syntactic Universal Dependencies (SUD), which avoid aggressive reconstruction of omitted structure \citep{gerdes2016sud}, and inference-based approaches that posit unpronounced but structurally recoverable material \citep{merchant2001syntax,hardt2004ellipsis}. In spontaneous bilingual conversation, constructions that represent disfluencies such as repetition and ellipsis are frequently discourse-dependent and structurally underspecified, leading to high annotation cost and lower attachment accuracy even for strong parsers. The continued degradation observed across systems therefore suggests that the challenge lies not only in model capacity or multilingual representation, but also in the representational assumptions underlying dependency annotation itself. Spoken conversational language often contains structures that are interactionally interpretable but only partially specified syntactically, making dependency reconstruction substantially more difficult than in canonical written language.

\begin {comment}

\paragraph{Results under Standard Metrics}
Table~\ref{tab:standard_metrics_main} reports performance under traditional attachment-based metrics (LAS, UAS, CLAS, UPOS-LAS) for three representative systems: a bilingual traditional parser, the LLM-based BiLingua pipeline, and the proposed DECAP framework. Across categories, all systems perform best on canonical sentences (``none'') and show substantial degradation on spoken-language phenomena such as repetition, ellipsis, and discourse-heavy constructions. While DECAP consistently outperforms both baselines under LAS and UPOS-LAS, for example, improving overall LAS from 0.31 (BiLingua) to 0.48 and UPOS-LAS from 0.70 to 0.87, standard metrics compress these gains and fail to distinguish linguistically principled analyses from benign structural variation. As a result, improvements in handling ambiguity and non-canonical structure are only partially reflected by attachment-based scores.

\paragraph{Results under FLEX-UD}
Table~\ref{tab:flexud_main} presents results under FLEX-UD for the same 3 systems. In contrast to standard metrics, FLEX-UD yields clearer separation across systems and sentence categories by explicitly accounting for ambiguity and error severity. DECAP achieves the highest overall FLEX-UD score (76.2), compared to 70.7 for BiLingua and 30.4 for the traditional parser, with especially large gains on repetition, ellipsis+, and discourse-heavy sentences. These improvements are driven not only by higher head and label accuracy, but by substantial reductions in catastrophic structural errors, which FLEX-UD penalizes more heavily than minor deviations. Additional results for monolingual and multilingual parsers are reported in Appendix~\ref{app:extendres}. Together, these findings demonstrate that FLEX-UD more faithfully captures qualitative improvements in spoken CSW parsing and that DECAP provides more robust and interpretable analyses than prior approaches.

\subsection{Discussion and Linguistic Analysis}
\label{sec:analysis}

\end{comment} 

\begin{comment}
 
\begin{figure}[t!]
    \centering
    \includegraphics[width=7cm]{images/upos-las.png}
   \caption{UPOS-LAS by Category and Parser; DECAP is the GPT-4.1 agent and performs the best across all categories.}
    \label{fig:upos-las}
\end{figure}

\begin{figure}[t!]
    \centering
    \includegraphics[width=7cm]{images/las.png}
   \caption{LAS by Category and Parser. DECAP is the GPT-4.1 agent and performs the best across all categories.}
    \label{fig:las}
\end{figure}   
\end{comment}

\begin {comment}
\paragraph{UPOS-LAS and LAS reflect complementary sources of difficulty.}
As shown in Figure~\ref{fig:upos-las}, UPOS-LAS exhibits a highly consistent hierarchy across bilingual and multilingual parsers, with repetition achieving the highest scores, followed by canonical sentences (``none''), ellipsis, and complex constructions. Averaged across systems, repetition reaches approximately 0.80-0.90 UPOS-LAS, compared to 0.75-0.85 for none, 0.60-0.70 for ellipsis, and 0.50-0.65 for complex sentences. This ordering reflects the sensitivity of UPOS-LAS to surface availability and local morphosyntactic cues: repetition maximizes overt lexical material, while ellipsis removes cues and complex constructions increase structural depth. Crucially, this hierarchy is stable across bilingual and multilingual parsers, whereas monolingual parsers score substantially lower, indicating that UPOS-LAS performance on CSW primarily reflects cross-lingual representational capacity rather than construction-specific heuristics. In contrast, LAS reveals a different hierarchy driven by structural ambiguity and recoverability.
\paragraph{LAS: structural ambiguity and recoverability.}
Insights from figure~\ref{fig:las} show that LAS yields a reversed but equally systematic hierarchy, with canonical and complex sentences achieving the highest accuracy, and ellipsis and repetition consistently performing worst (none \(\approx\) complex \(>\) ellipsis \(\approx\) repetition). Averaged across parsers, none and complex constructions reach approximately 0.75-0.85 LAS, while ellipsis drops to 0.55-0.70 and repetition to 0.50-0.65. This pattern aligns with established findings that ellipsis requires recovery of unexpressed structure and discourse-level inference \citep{merchant2001syntax, dobrovoljc2022ellipsis, cavar2024spoken}, a difficulty corroborated by annotation effort: sentences with ellipsis required roughly two times as much annotation time as sentences without ellipsis. Repetition also reduces LAS by introducing competing attachment sites despite surface redundancy. LLM-based parsers follow this hierarchy but exhibit sharper declines for ellipsis and repetition, whereas the traditional bilingual parser degrades more uniformly, suggesting differences in sensitivity to structural ambiguity rather than some random errors.
These construction-specific difficulties are systematically amplified in bilingual production, where repetition, reformulation, and ellipsis arise more; a detailed analysis of bilingualism as a complexity multiplier is provided in Appendix~\ref{app:extended_analysis}.
\end{comment}

%% file: 07_conclusion.tex
\section{Conclusion}
\label{sec:conc}

Spoken bilingual conversational language exposes a persistent mismatch between written-language parsing assumptions and the structural variability of interactional speech. To address these challenges, we introduced a linguistically grounded taxonomy of conversational bilingual phenomena, \textbf{SpokeBench}, an expert-annotated benchmark for spoken conversational parsing, \textbf{Flex-UD}, an ambiguity-aware evaluation metric designed for structurally variable conversational language, and \textbf{DECAP}, a modular agentic parsing framework that separates spoken-phenomena handling from core syntactic analysis. Experiments across both proprietary and open-weight LLMs demonstrate substantial improvements in robustness to conversational phenomena, while Flex-UD reveals meaningful structural consistency and parser differences that remain partially hidden under strict attachment-based evaluation. Together, these findings suggest that progress in conversational bilingual parsing depends not only on stronger models but also on evaluation and annotation frameworks that better account for ambiguity, recoverability, and the discourse-driven nature of spontaneous speech.

\section*{Limitations}
\label{sec:lim}

This work focuses on depth and linguistic complexity rather than scale. \textbf{SpokeBench} is intentionally small and curated, prioritizing expert annotation and challenging Spoken Bilingual Conversation phenomena over broad coverage across speakers, dialects, or interactional contexts. While this design enables detailed analysis, extending the benchmark to larger and more diverse spoken corpora would strengthen generalizability. In addition, our experiments focus on English-Spanish code-switching; although the identified phenomena are common in many bilingual settings, their distribution and interaction may vary across language pairs and sociolinguistic environments. Finally, we proposed an instruction-grounded framework rather than a trained parser, which enables extensibility without retraining, but is comparatively slower and depends on the quality of prompts and the underlying language models. Exploring hybrid approaches that combine spoken-aware system design with learned representations remains an important avenue for future research.

\section*{Ethical Considerations}
This work uses data from the Miami Corpus of English-Spanish code-switching \citep{deuchar2014building}, which was collected and released for research purposes; no new data were collected and no personally identifiable information was accessed. While our methods aim to improve parsing and evaluation of spoken bilingual language, LLM-based annotation may propagate biases or misrepresent code-switching practices if applied without linguistic oversight, and should therefore be used cautiously outside research settings. We used AI-assisted tools (e.g., ChatGPT and Grammarly) solely for grammatical refinement and clarity of presentation.

%% file: 08_appendix.tex
\appendix
\section{Distribution of SpokeBench}
\label{app:spokebench}
Table \ref{tab:spokebench_subset} reports the distribution of sentences in \textbf{SpokeBench} by Spoken Bilingual Conversation phenomenon. The benchmark was designed to balance coverage of frequent spoken phenomena with a controlled set of highly complex constructions, enabling systematic evaluation of parsers under varied spoken-language conditions.

\begin{table}[htbp]
\centering
\small
\renewcommand{\arraystretch}{1.2}
\begin{tabular}{l r}
\hline
\textbf{Category} & \textbf{\# Sentences} \\
\hline
Repetition & 10 \\
Repetition+ & 20 \\
Contractions English (EN) & 10 \\
Contractions Spanish (ES) & 10 \\
Ellipsis & 10 \\
Ellipsis+ & 20 \\
Discourse & 10 \\
Discourse+ & 20 \\
Fillers & 10 \\
Complex (3+ phenomena) & 20 \\
Others  & 20 \\
\hline
\textbf{Total} & \textbf{160} \\
\hline
\end{tabular}
\caption{Composition of the SpokeBench benchmark by Spoken Bilingual Conversation phenomenon.}
\label{tab:spokebench_subset}
\end{table}

\section{Examples of Spoken Bilingual Conversation Phenomena}
\label{app:examples}
Table \ref{tab:appendix_examples} provides representative examples of the spoken code-switching phenomena included in our taxonomy, drawn from the Miami Corpus. Each example illustrates a characteristic non-canonical structure encountered in conversational bilingual speech, along with an English translation to support interpretability for non-Spanish speakers.

\begin{table*}[t]
\centering
\footnotesize
\renewcommand{\arraystretch}{1.3}
\begin{tabular}{p{0.18\textwidth} p{0.36\textwidth} p{0.36\textwidth}}
\hline
\textbf{Category} & \textbf{Example (Miami Corpus)} & \textbf{English Translation} \\
\hline
Repetition &
entonces then I will I will I will not be buying any stuff for you this weekend . &
So then I will, I will, I will not be buying any stuff for you this weekend. \\

Discourse Elements &
uhhuh bueno es verdad . &
Uh-huh, well, it’s true. \\

Ellipsis &
pero si they are covered from . &
But if they are covered from [it]. \\

Contractions &
and tú sabes it wasn't the same . &
And you know it wasn’t the same. \\

Compound Words &
cuando uno va al swimming pool ahí no me gusta . &
When one goes to the swimming pool, I don’t like it there. \\

False Start &
you have you gotta show tu tía . &
You have.. you gotta show your aunt. \\

Filler Words &
fíjate que they gave them an honorary uh diploma . &
Look, they gave them an honorary, uh, diploma. \\

Slang / Curse Words &
I mean I'm thinking coño if I'm not here what the hell would happen to me ? &
I mean, I’m thinking, damn, if I’m not here, what the hell would happen to me? \\

Enclisis &
more jugo dile a Carla que te dé more jugo &
More juice, tell Carla to give you more juice. \\
\hline
\end{tabular}
\caption{Representative examples of some spoken bilingual conversational language phenomena from the Miami Corpus, with English translations.}
\label{tab:appendix_examples}
\end{table*}

\section{Condensed Annotation Guidelines}
\label{app:guidelines}
\subsection{General Annotation Principles}
Annotations in SpokeBench follow UD conventions with targeted extensions for spoken, disfluent, and code-switched data, guided by the principles below:
\begin{itemize}[topsep=2pt, itemsep=1pt, parsep=0pt, partopsep=0pt]
    \item \textbf{UD-first consistency:} Annotators adhered to standard UD conventions unless a spoken-language phenomenon required an explicit, documented deviation.
    
    \item \textbf{Single-root constraint:} Every sentence was required to contain exactly one syntactic root, including minimal, elliptical, or fragmentary utterances (e.g., \textit{si mmh}). The root was assigned to the lexical item carrying the greatest semantic weight.
    
    \item \textbf{Ambiguity tolerance:} Structural ambiguity was treated as an inherent property of spoken conversational bilingual data rather than as an annotation error. Annotators were instructed to prefer linguistically plausible analyses over maximally specific ones.
    
    \item \textbf{Internal consistency:} All annotations were reviewed for internal consistency, with particular attention to agreement between UPOS tags and dependency relations.
    
    \item \textbf{Conservative tokenization:} Original corpus tokenization was preserved whenever possible. Token splitting or insertion of additional rows was permitted only when necessary to support accurate syntactic analysis.
\end{itemize}

\subsection{Treatment of Spoken-Language Phenomena}

This subsection summarizes the core annotation rules applied to recurring spoken-language phenomena in SpokeBench. These rules were designed to minimize speculative reconstruction while ensuring consistent and transparent treatment across annotators.

\paragraph{Repetition and Repairs}
Repeated tokens that function as speech repairs or disfluencies were annotated using the dependency relation \texttt{reparandum}. Annotators were instructed to identify the intended syntactic structure and attach repeated or overridden material as dependents of the corresponding head in the corrected structure. This procedure was applied uniformly to both monolingual and code-switched repetitions.

\paragraph{Contractions}
English and Spanish contractions (e.g., \textit{wasn't}, \textit{del}) were split into their component morphemes by inserting additional tokens immediately following the original form. The expanded tokens were then annotated according to standard UD conventions. This approach preserves morphological transparency while enabling accurate dependency assignment.

\paragraph{Multiword Expressions and Compounds}
Fixed expressions and conventionalized compounds (e.g., \textit{you know}, \textit{a lot}) were treated as single syntactic units rather than annotated compositionally. Annotators combined such expressions using underscore-separated tokens and assigned UPOS and dependency relations corresponding to their discourse or syntactic function.

\paragraph{Ellipsis and Fragmented Utterances}
Elliptical or truncated constructions were handled according to the recoverability of the intended meaning. Tokens that could be removed without altering the core interpretation of the utterance were labeled as \texttt{reparandum}. Tokens whose syntactic heads were missing and could not be confidently reconstructed were assigned the relation \texttt{dep} and attached to the sentence root. In fragmentary utterances lacking an explicit predicate, the most semantically salient token was designated as the root.

\paragraph{Discourse Markers and Fillers}
Discourse markers, interjections, and hesitation sounds (e.g., \textit{uh}, \textit{mmh}) were annotated with \texttt{UPOS=INTJ} and assigned the dependency relation \texttt{discourse}. Fillers were treated as a distinct subclass to facilitate later analysis but followed the same structural annotation principles.

\subsection{Review and Adjudication Procedure}
Each sentence in SpokeBench was independently annotated by at least two annotators. Annotations were subsequently reviewed using a graded acceptability scale that distinguished between acceptable variation and substantive errors. Reviewers were instructed to assess whether an analysis was linguistically defensible under UD rather than whether it matched their personal preference. Sentences flagged as borderline or unacceptable were escalated for expert discussion. Final annotations were determined through deliberation involving senior linguists on the project, ensuring consistency across the benchmark. This process resulted in a gold-standard dataset that reflects informed consensus while acknowledging the inherent ambiguity of spoken bilingual conversations.

\section{DECAP Interaction Protocol}
\label{app:int}
The details of the DECAP agent workflow are shown below, in Algorithm \ref{alg:decap}.

\begin{algorithm}[htbp]
\caption{DECAP per-sentence interaction rule}
\label{alg:decap}
\small
\begin{algorithmic}[1]
\REQUIRE Tokenized sentence $S = \{(t_i, \text{lang}_i)\}_{i=1}^n$
\ENSURE UD parse with validated structure and confidence annotations

\STATE $\mathcal{S} \leftarrow \textsc{SPH}(S)$ 
\COMMENT{Detect spoken phenomena; propose tokenization edits}

\STATE $\mathcal{L} \leftarrow \textsc{LSR}(\mathcal{S})$ 
\COMMENT{Apply language-specific normalization and MWEs}

\STATE $\mathcal{C} \leftarrow \textsc{Core}(\mathcal{L})$ 
\COMMENT{Assign UPOS, heads, and dependency relations}

\STATE $\mathcal{F} \leftarrow \textsc{V/R}(\mathcal{S}, \mathcal{L}, \mathcal{C})$ 
\COMMENT{Enforce single root, acyclicity, and consistency}

\STATE $\mathcal{F}$ includes confidence scores and logs any structural repair

\RETURN $\mathcal{F}$
\end{algorithmic}
\end{algorithm}

\section{FLEX-UD Severity Examples}
\label{app:flexud_examples}
\paragraph{Catastrophic Errors.}
Catastrophic errors correspond to violations that undermine the global structural validity or interpretability of a parse. These include cases where a dotted multiword expression (MWE) is required by the gold annotation but omitted by the system; reparandum tokens attached to unrelated subtrees, resulting in incorrect root selection or major structural distortion; and invalid head references that persist after canonicalization. Such errors receive large penalty contributions in Flex-UD, typically in the range of \(P = 0.25\) to \(0.6\) per issue (clipped), reflecting their disproportionate impact on parse quality.

\paragraph{Minor Errors.}
Minor errors correspond to linguistically plausible deviations that do not compromise the overall structure of the parse. Examples include tolerant POS substitutions (e.g., VERB\(\leftrightarrow\)AUX), near-miss dependency relations (e.g., \texttt{obj}\(\leftrightarrow\)\texttt{obl}), and other UPOS or DEPREL mismatches that fall within predefined closeness classes. These errors incur small penalty contributions, typically in the range of \(P = 0.01\) to \(0.05\) per issue, allowing Flex-UD to distinguish acceptable variation from substantive failure.

\section{Prompts Used for DECAP}
\label{app:decapprompt}
The prompt structures for the 4 agents used in the DECAP pipeline are described below. Each prompt corresponds to a dedicated agent with a narrowly scoped responsibility in the annotation process, progressing from spoken-phenomena detection to language-specific normalization, core UD assignment, and final verification and ranking.

\clearpage

\begin{tcolorbox}[
  colback=white,
  colframe=gray!60!black,
  width=\textwidth,
  boxrule=0.5pt,
  arc=2pt,
  left=6pt,
  right=6pt,
  top=6pt,
  bottom=6pt,
  breakable,
  title=\textbf{Prompt for Spoken-Phenomena Handler (SPH)},
  fonttitle=\bfseries,
  coltitle=white
]
\textbf{System Prompt (SPH).} \\
You are the \textbf{Spoken-Phenomena Handler (SPH)} for Spanish--English conversational data. Your responsibility is to detect spoken-language phenomena, propose minimal tokenization edits (e.g., contraction splits and multiword expressions), and produce a single structured JSON object for downstream agents. Return \emph{only} valid JSON and behave deterministically.

\vspace{0.5em}
\textbf{User Prompt (SPH).} \\
\textbf{Input:} A JSON array of tokens representing one sentence, including sentence ID, token index, surface form, and language tag.

\textbf{Goal:} Produce a validated SPH JSON object that:
\begin{itemize}[leftmargin=1.2em, itemsep=0.2em, topsep=0pt]
  \item identifies spoken-language phenomena (e.g., repetition, ellipsis, discourse, fillers),
  \item proposes necessary tokenization edits (contractions, enclisis, MWEs),
  \item constructs a complete and consistent proposed ID mapping, and
  \item reports a brief summary and confidence score.
\end{itemize}

\vspace{0.5em}
\textbf{Key Rules (excerpt):}
\begin{itemize}[leftmargin=1.2em, itemsep=0.2em, topsep=0pt]
  \item Contractions and enclitic forms are split only when they correspond to multiple UD nodes.
  \item Fixed multiword expressions are collapsed into dotted nodes when they function as a single syntactic unit.
  \item Repetitions and repairs are marked as \texttt{reparandum} and linked to the intended head.
  \item Elliptical or stranded tokens are labeled conservatively and anchored to the root when necessary.
\end{itemize}

Additional rules governing ID assignment, mandatory MWEs, validation checks, and output schema are provided in the full prompt specification in the GitHub (\ldots).

\vspace{0.5em}
\textbf{Output:} A single JSON object containing normalized tokens, spoken-phenomena labels, a proposed ID map, confidence scores, and brief summary notes. No explanatory text is permitted.
\end{tcolorbox}

% Full-width, float-page tcolorbox (LSR)
\begin{tcolorbox}[
  enhanced,
  float*=p,                     % starred float -> spans both columns, placed on float page
  colback=gray!5,
  colframe=gray!60!black,
  width=\textwidth,
  boxrule=0.5pt,
  arc=2pt,
  left=6pt, right=6pt, top=6pt, bottom=6pt,
  breakable,
  title=\textbf{Prompt for Language-Specific Resolver (LSR)},
  fonttitle=\bfseries,
  coltitle=white
]
% <--- (content unchanged) --->
\textbf{System Prompt (LSR).} \\
You are the \textbf{Language-Specific Resolver (LSR)} for Spanish--English code-switched conversational data. Your single responsibility is to accept the SPH JSON and apply conservative language-specific normalization (contraction/enclisis expansion, MWE confirmation/creation, lemma suggestions, and language-confidence scores). Respect SPH edits; return \emph{only} one JSON object and behave deterministically.

\vspace{0.5em}
\textbf{Input:} The SPH JSON object with keys including \texttt{sentence\_id}, \texttt{original\_tokens}, \texttt{tokens}, and \texttt{proposed\_id\_map}.

\textbf{Goal:} Produce an updated JSON that:
\begin{itemize}[leftmargin=1.2em, itemsep=0em, topsep=0pt]
  \item confirms or conservatively expands standard contractions/enclitics (INTEGER-SHIFT),
  \item confirms or (rarely) creates high-precision MWEs as dotted nodes (whitelist + conservative rule),
  \item supplies lowercase \texttt{lemma} suggestions and \texttt{lsr\_confidence} per token,
  \item updates \texttt{proposed\_id\_map} consistently for any insertions.
\end{itemize}

\vspace{0.5em}
\textbf{Key Rules (excerpt):}
\begin{itemize}[leftmargin=1.2em, itemsep=0.2em, topsep=0pt]
  \item \textbf{Respect SPH authority}: do not undo SPH tokenization edits; only add splits when SPH omitted a clear, standard contraction/enclitic.
  \item \textbf{Contraction handling (INTEGER-SHIFT)}: expand only standard contractions that map to separate UD nodes; insert new integer proposed\_IDs and shift subsequent IDs +1, reflecting changes in \texttt{proposed\_id\_map}.
  \item \textbf{MWE creation (DOTTED IDs)}: auto-combine mandatory whitelist MWEs (case-insensitive). Create other MWEs only under conservative, high-confidence criteria; add dotted node \texttt{<start>.1} placed after the span.
  \item \textbf{Lemmas \& language confidence}: provide best-effort lemmas (verbs → infinitive, nouns → singular) or \texttt{null}; set \texttt{lsr\_confidence} in [0.0,1.0] and one-line \texttt{lsr\_notes} for nontrivial edits.
  \item \textbf{IDs and validation}: do not change original \texttt{token\_index}; ensure unique proposed\_IDs, complete \texttt{proposed\_id\_map}, and consistency after integer-shifts.
\end{itemize}

\vspace{0.5em}
\textbf{Examples (excerpt):}
\begin{itemize}[leftmargin=1.2em, itemsep=0.2em, topsep=0pt]
  \item \textbf{Contraction split}: \texttt{"don't"} → \texttt{"do"} (orig id) + inserted \texttt{"not"} (new id); update map so orig id maps to \texttt{["5","6"]}.
  \item \textbf{MWE (whitelist)}: \texttt{"pitta bread"} → keep 6,7 integer rows and add dotted 6.1 with \texttt{split\_token="pitta\_bread"}; reflect in map.
\end{itemize}

Additional details on whitelist entries, edge cases, full schema, and validations are available in the full prompt spec on the project repository (\ldots).

\vspace{0.5em}
\textbf{Output:} A single JSON object with keys \texttt{"sentence\_id"}, \texttt{"tokens"} (including \texttt{proposed\_ID}, \texttt{split\_token}, \texttt{lang\_tag}, \texttt{lemma}, \texttt{lsr\_notes}, \texttt{lsr\_confidence}, \texttt{mwe}), \texttt{"proposed\_id\_map"}, and \texttt{"summary\_notes"}. No additional text is permitted.
\end{tcolorbox}

% Full-width, float-page tcolorbox (Core)
\begin{tcolorbox}[
  enhanced,
  float*=p,                     % starred float -> spans both columns, placed on float page
  colback=gray!5,
  colframe=gray!60!black,
  width=\textwidth,
  boxrule=0.5pt,
  arc=2pt,
  left=6pt, right=6pt, top=6pt, bottom=6pt,
  breakable,
  before=\vspace{0pt},
  after=\vspace{0pt},
  title=\textbf{Prompt for Core UD Assigner (Core)},
  fonttitle=\bfseries,
  coltitle=white
]
\textbf{System Prompt (Core).} \\
You are the \textbf{Core UD Assigner} for Spanish--English conversational data. Your task is to assign UD-style annotations (UPOS, HEAD\_ID, DEPREL) following the Miami Gold Subset spoken-language rules. Respect tokenization and proposed\_IDs from upstream (SPH + LSR) and spoken-language directives (spoken\_label / spoken\_anchor). Annotate only dotted MWE nodes (n.1); leave MWE component integer rows unannotated. Ensure exactly one root. Return \emph{only} a single JSON object (no commentary).

\vspace{0.5em}
\textbf{User Prompt (Core).} \\
\textbf{Input:} LSR JSON object with keys: \texttt{sentence\_id, tokens (proposed\_ID, split\_token, lang\_tag, lemma, lsr\_confidence, mwe, spoken\_label, spoken\_anchor)}, and \texttt{proposed\_id\_map}.\\
\textbf{Goal:} Produce an annotated JSON with UD fields for every annotatable node, obeying spoken-label mappings and UD constraints.

\vspace{0.5em}
\textbf{Key Principles (high-level):}
\begin{itemize}[leftmargin=1.2em, itemsep=0.2em, topsep=0pt]
  \item \textbf{Respect upstream}: follow SPH/LSR spoken\_label and recommended head (spoken\_anchor) when provided.
  \item \textbf{Spoken-label $\to$ UD mapping}: e.g., \texttt{reparandum} $\to$ DEPREL=\texttt{rep}; \texttt{dep} $\to$ DEPREL=\texttt{dep}; \texttt{discourse/filler} $\to$ UPOS=\texttt{INTJ} and DEPREL=\texttt{discourse} (attach to root unless syntactically integrated).
  \item \textbf{Single root}: exactly one token with \texttt{HEAD\_ID="0"} and \texttt{DEPREL="root"}; prefer finite VERB, else central NOUN/PRON, else communicative token.
  \item \textbf{Allowed tag sets}: UPOS and DEPREL must be chosen from the Miami-approved lists (use only allowed values).
\end{itemize}

\vspace{0.4em}
\textbf{Output Schema (required, single JSON object):}
\begin{verbatim}
{
 "sentence_id": "<id>",
 "annotated_tokens": [
   {
     "proposed_ID": "<str>",
     "FORM": "<split_token or empty>",
     "LEMMA": "<lemma or empty>",
     "UPOS": "<allowed UPOS or empty>",
     "HEAD_ID": "<proposed_ID or '0' or empty>",
     "HEAD_FORM": "<FORM of head or 'root' or empty>",
     "DEPREL": "<allowed DEPREL or empty>",
     "core_confidence": 0.0-1.0,
     "core_notes": "<one-line justification>"
   }, ...
 ],
 "summary_notes": "<one-line summary>"
}
\end{verbatim}

\vspace{0.4em}
\textbf{Essential validations (must pass):}
...

\vspace{0.4em}
\textbf{Return:} exactly one JSON object following the schema above. Do not include any additional text.
\end{tcolorbox}

% Full-width, float-page tcolorbox (Core)
\begin{tcolorbox}[
  enhanced,
  float*=p,                     % starred float -> spans both columns, placed on float page
  colback=gray!5,
  colframe=gray!60!black,
  width=\textwidth,
  boxrule=0.5pt,
  arc=2pt,
  left=6pt, right=6pt, top=6pt, bottom=6pt,
  breakable,
  before=\vspace{0pt},
  after=\vspace{0pt},
  title=\textbf{Prompt for Verifier \& Ranker (V/R)},
  fonttitle=\bfseries,
  coltitle=white
]\small
\textbf{System Prompt (V/R).} \\
You are the Verifier \& Ranker. Merge SPH, LSR and Core outputs for one sentence; enforce ID/HEAD consistency; repair structural issues (single root, cycles); compute per-token \texttt{final\_confidence} and \texttt{penalty}; remap proposed\_IDs to sheet integers when requested; and emit sheet-ready rows plus an adjudication log. Respect authoritative ordering: Core $\rhd$ LSR $\rhd$ SPH. Return \emph{only} the JSON described below.
\vspace{0.5em}
\textbf{User Prompt (v/R).} \\
\textbf{Input \& Goal.} \\
Input: JSON bundle with keys \texttt{"sph"}, \texttt{"lsr"}, \texttt{"core"}. Goal: produce validated final token table (all integer rows, inserted integers, dotted MWE nodes), mapping to sheet IDs, plus \texttt{adjudication\_log} and a one-line \texttt{final\_summary}.

\vspace{0.5em}
\textbf{Condensed procedure (deterministic):}
\begin{itemize}[leftmargin=1.2em,itemsep=0em,topsep=0pt]
  \item Canonicalize nodes from SPH original order, inserting LSR/SPH-added integers and dotted MWEs.
  \item Merge annotations (prefer Core; fill from LSR then SPH).
  \item Validate/remap HEADs (HEAD\_FORM match, then SPH spoken\_anchor, else attach to root) and log.
  \item Enforce single root by priority (finite VERB, AUX, central NOUN/PRON, else highest confidence) and log.
  \item Detect \& repair cycles using combined\_conf; reattach lowest node(s) to root until acyclic; log numeric rationale.
  \item Compute final\_confidence and penalty (per rules) and add adjudication notes.
  \item Remap to sheet\_IDs (1..N) and produce sheet\_HEAD\_ID mapping.
  \item Emit final JSON and validate
\end{itemize}

\vspace{0.5em}
\textbf{Required output shape (exact):}
\begin{verbatim}
{
 "sentence_id":"<id>",
 "final_tokens":[
   {
     "sentence_id":<id>,
     "orig_token_index":<int>,
     "split_token":"<str>",
     "ID":"<proposed_ID>",
     "sheet_ID":<int>,
     "FORM":"<str or blank>",
     "LEMMA":"<str or blank>",
     "UPOS":"<or blank>",
     "HEAD_ID":"<proposed_ID or '0'>",
     "sheet_HEAD_ID":<int or 0>,
     "HEAD":"<HEAD_FORM or blank>",
     "DEPREL":"<or blank>",
     "final_confidence":0.0-1.0,
     "penalty":0.0-1.0,
     "adjudication_note":"<short>"
   },
   ...
 ],
 "adjudication_log":[ "...", "..." ],
 "final_summary":"<one-line>"
}
\end{verbatim}

\vspace{0.4em}
\textbf{Determinism \& logging:} use deterministic tie-breakers (prefer lower sheet\_ID). Each structural fix must add one bullet to \texttt{adjudication\_log} stating what changed, why (include numeric confidences), and which signals supported the decision.

\end{tcolorbox}

%% file: 00_acl_latex.bbl
\begin{thebibliography}{28}
\providecommand{\natexlab}[1]{#1}

\bibitem[{Bai et~al.(2025)Bai, Wu, Chen, Wang, Chen, Zhang, and Zhang}]{bai2025constituency}
Xuefeng Bai, Jialong Wu, Yulong Chen, Zhongqing Wang, Kehai Chen, Min Zhang, and Yue Zhang. 2025.
\newblock \href {https://doi.org/10.1109/TASLPRO.2025.3600867} {Constituency parsing using llms}.
\newblock \emph{IEEE Transactions on Audio, Speech and Language Processing}, 33:3762--3775.
\newblock Vol. 33.

\bibitem[{Bhat et~al.(2018)Bhat, Bhat, Shrivastava, and Sharma}]{bhat2018universal}
Irshad Bhat, Riyaz~Ahmad Bhat, Manish Shrivastava, and Dipti~Misra Sharma. 2018.
\newblock Universal dependency parsing for hindi-english code-switching.
\newblock In \emph{Proceedings of the 2018 Conference of the North American Chapter of the Association for Computational Linguistics: Human Language Technologies, Volume 1 (Long Papers)}, pages 987--998.

\bibitem[{{\c{C}}etino{\u{g}}lu and {\c{C}}{\"o}ltekin(2019)}]{ccetinouglu2019challenges}
{\"O}zlem {\c{C}}etino{\u{g}}lu and {\c{C}}a{\u{g}}r{\i} {\c{C}}{\"o}ltekin. 2019.
\newblock Challenges of annotating a code-switching treebank.
\newblock In \emph{Proceedings of the 18th international workshop on Treebanks and Linguistic Theories (TLT, SyntaxFest 2019)}, pages 82--90.

\bibitem[{{\c{C}}etino{\u{g}}lu and {\c{C}}{\"o}ltekin(2023)}]{ccetinouglu2023two}
{\"O}zlem {\c{C}}etino{\u{g}}lu and {\c{C}}a{\u{g}}r{\i} {\c{C}}{\"o}ltekin. 2023.
\newblock Two languages, one treebank: building a turkish--german code-switching treebank and its challenges.
\newblock \emph{Language Resources and Evaluation}, 57(2):545--579.

\bibitem[{De~Leon et~al.(2024)De~Leon, Madabushi, and Lee}]{de2024code}
Frances Adriana~Laureano De~Leon, Harish~Tayyar Madabushi, and Mark Lee. 2024.
\newblock Code-mixed probes show how pre-trained models generalise on code-switched text.
\newblock In \emph{Proceedings of the 2024 Joint International Conference on Computational Linguistics, Language Resources and Evaluation (LREC-COLING 2024)}, pages 3457--3468.

\bibitem[{de~Marneffe et~al.(2021)de~Marneffe, Manning, Nivre, and Zeman}]{deMarneffe2021UD}
Marie-Catherine de~Marneffe, Christopher~D. Manning, Joakim Nivre, and Daniel Zeman. 2021.
\newblock \href {https://doi.org/10.1162/coli_a_00402} {Universal dependencies}.
\newblock \emph{Computational Linguistics}, 47(2):255--308.

\bibitem[{Deuchar et~al.(2014)Deuchar, Davies, Herring, Parafita~Couto, and Carter}]{deuchar2014building}
Margaret Deuchar, Peter Davies, Judith Herring, María~C. Parafita~Couto, and Dan Carter. 2014.
\newblock Building bilingual corpora.
\newblock In Enlli~M. Thomas and Ineke Mennen, editors, \emph{Advances in the Study of Bilingualism}, pages 93--110. Multilingual Matters, Bristol.

\bibitem[{Dobrovoljc(2022)}]{dobrovoljc2022spoken}
Kaja Dobrovoljc. 2022.
\newblock Spoken language treebanks in universal dependencies: An overview.
\newblock In \emph{Proceedings of the Thirteenth Language Resources and Evaluation Conference}, pages 1798--1806.

\bibitem[{Gerdes et~al.(2019{\natexlab{a}})Gerdes, Guillaume, Kahane, and Perrier}]{gerdes2016sud}
Kim Gerdes, Bruno Guillaume, Sylvain Kahane, and Guy Perrier. 2019{\natexlab{a}}.
\newblock Improving surface-syntactic universal dependencies (sud): Mwes and deep syntactic features.
\newblock In \emph{Proceedings of the 18th international workshop on treebanks and linguistic theories (TLT, SyntaxFest 2019)}, pages 126--132.

\bibitem[{Gerdes et~al.(2019{\natexlab{b}})Gerdes, Guillaume, Kahane, and Perrier}]{gerdes2019improving}
Kim Gerdes, Bruno Guillaume, Sylvain Kahane, and Guy Perrier. 2019{\natexlab{b}}.
\newblock Improving surface-syntactic universal dependencies (sud): surface-syntactic relations and deep syntactic features.
\newblock In \emph{TLT 2019-18th International Workshop on Treebanks and Linguistic Theories}, pages 126--132. Association for Computational Linguistics.

\bibitem[{G{\'o}mez-Rodr{\'\i}guez et~al.(2024)G{\'o}mez-Rodr{\'\i}guez, Imran, Vilares, Solera, and Kellert}]{gomez2024dancing}
Carlos G{\'o}mez-Rodr{\'\i}guez, Muhammad Imran, David Vilares, Elena Solera, and Olga Kellert. 2024.
\newblock Dancing in the syntax forest: fast, accurate and explainable sentiment analysis with salsa.
\newblock In \emph{SEPLN–CEDI-PD 2024. Seminar of the Spanish Society for Natural Language Processing: Projects and System Demonstrations}, volume 3729 of \emph{CEUR Workshop Proceedings}, pages 12--17, A Coru{\~n}a, Spain.

\bibitem[{Hardt and Romero(2004)}]{hardt2004ellipsis}
Daniel Hardt and Maribel Romero. 2004.
\newblock Ellipsis and the structure of discourse.
\newblock In \emph{Proceedings of SALT 14}.

\bibitem[{Huzaifah et~al.(2024)Huzaifah, Zheng, Chanpaisit, and Wu}]{huzaifah2024evaluating}
Muhammad Huzaifah, Weihua Zheng, Nattapol Chanpaisit, and Kui Wu. 2024.
\newblock Evaluating code-switching translation with large language models.
\newblock In \emph{Proceedings of the 2024 Joint International Conference on Computational Linguistics, Language Resources and Evaluation (LREC-COLING 2024)}, pages 6381--6394.

\bibitem[{Imran et~al.(2026{\natexlab{a}})Imran, Kellert, and G{\'o}mez-Rodr{\'\i}guez}]{imran2024syntax}
Muhammad Imran, Olga Kellert, and Carlos G{\'o}mez-Rodr{\'\i}guez. 2026{\natexlab{a}}.
\newblock \href {https://doi.org/10.7717/peerj-cs.3519} {A syntax-injected approach for faster and more accurate sentiment analysis}.
\newblock \emph{PeerJ Computer Science}, 12:e3519.

\bibitem[{Imran et~al.(2026{\natexlab{b}})Imran, Zamaraeva, and G{\'o}mez-Rodr{\'\i}guez}]{imran2025synner}
Muhammad Imran, Olga Zamaraeva, and Carlos G{\'o}mez-Rodr{\'\i}guez. 2026{\natexlab{b}}.
\newblock \href {https://doi.org/10.1093/jamiaopen/ooaf149} {Synner: Syntax-infused named entity recognition in the biomedical domain}.
\newblock \emph{JAMIA Open}, 9(1):ooaf149.

\bibitem[{James et~al.(2022)James, Yogarajan, Shields, Watson, Keegan, Mahelona, and Jones}]{james2022language}
Jesin James, Vithya Yogarajan, Isabella Shields, Catherine~I Watson, Peter Keegan, Keoni Mahelona, and Peter-Lucas Jones. 2022.
\newblock Language models for code-switch detection of te reo m{\=a}ori and english in a low-resource setting.
\newblock In \emph{Findings of the Association for Computational Linguistics: NAACL 2022}, pages 650--660.

\bibitem[{Kahane et~al.(2021)Kahane, Caron, Strickland, and Gerdes}]{kahane2021annotation}
Sylvain Kahane, Bernard Caron, Emmett Strickland, and Kim Gerdes. 2021.
\newblock Annotation guidelines of ud and sud treebanks for spoken corpora.
\newblock In \emph{Proceedings of the 20th International Workshop on Treebanks and Linguistic Theories (TLT, SyntaxFest 2021)}, pages pp--35. Association for Computational Linguistics.

\bibitem[{Kellert et~al.(2025)Kellert, Tyagi, Imran, Licona-Guevara, and G{\'o}mez-Rodr{\'i}guez}]{kellert2025parsing}
Olga Kellert, Nemika Tyagi, Muhammad Imran, Nelvin Licona-Guevara, and Carlos G{\'o}mez-Rodr{\'i}guez. 2025.
\newblock Parsing the switch: {LLM}-based universal dependency annotation for code-switched language.
\newblock In \emph{Findings of the Association for Computational Linguistics: EMNLP 2025}, pages 15934--15949, Suzhou, China. Association for Computational Linguistics.

\bibitem[{Kyle et~al.(2022)Kyle, Eguchi, Miller, and Sither}]{kyle2022dependency}
Kristopher Kyle, Masaki Eguchi, Aaron Miller, and Theodore Sither. 2022.
\newblock A dependency treebank of spoken second language english.
\newblock In \emph{Proceedings of the 17th workshop on innovative use of NLP for building educational applications (BEA 2022)}, pages 39--45.

\bibitem[{Merchant(2001)}]{merchant2001syntax}
Jason Merchant. 2001.
\newblock \emph{The Syntax of Silence: Sluicing, Islands, and the Theory of Ellipsis}.
\newblock Oxford University Press.

\bibitem[{Nguyen et~al.(2021)Nguyen, Lai, Veyseh, and Nguyen}]{nguyen2021trankit}
Minh~Van Nguyen, Viet~Dac Lai, Amir Pouran~Ben Veyseh, and Thien~Huu Nguyen. 2021.
\newblock Trankit: A light-weight transformer-based toolkit for multilingual natural language processing.
\newblock In \emph{Proceedings of the 16th Conference of the European Chapter of the Association for Computational Linguistics: System Demonstrations}, pages 80--90. Association for Computational Linguistics.

\bibitem[{Nivre et~al.(2016)Nivre, de~Marneffe, Ginter, Goldberg, Haji{\v{c}}, Manning, McDonald, Petrov, Pyysalo, Silveira, Tsarfaty, and Zeman}]{nivre2016ud}
Joakim Nivre, Marie-Catherine de~Marneffe, Filip Ginter, Yoav Goldberg, Jan Haji{\v{c}}, Christopher~D. Manning, Ryan McDonald, Slav Petrov, Sampo Pyysalo, Natalia Silveira, Reut Tsarfaty, and Daniel Zeman. 2016.
\newblock Universal dependencies v1: A multilingual treebank collection.
\newblock In \emph{Proceedings of the Tenth International Conference on Language Resources and Evaluation (LREC'16)}, pages 1659--1666. European Language Resources Association (ELRA).

\bibitem[{Pulido et~al.(2025)Pulido, Pugh, and Liu}]{pulido2025speak}
Emiliana Pulido, Robert Pugh, and Zoey Liu. 2025.
\newblock I speak for the {\'a}rboles: Developing a dependency treebank for spanish l2 and heritage speakers.
\newblock In \emph{Proceedings of the 63rd Annual Meeting of the Association for Computational Linguistics (Volume 4: Student Research Workshop)}, pages 814--822.

\bibitem[{Sanguinetti et~al.(2023)Sanguinetti, Bosco, Cassidy, {\c{C}}etino{\u{g}}lu, Cignarella, Lynn, Rehbein, Ruppenhofer, Seddah, and Zeldes}]{sanguinetti2023treebanking}
Manuela Sanguinetti, Cristina Bosco, Lauren Cassidy, {\"O}zlem {\c{C}}etino{\u{g}}lu, Alessandra~Teresa Cignarella, Teresa Lynn, Ines Rehbein, Josef Ruppenhofer, Djam{\'e} Seddah, and Amir Zeldes. 2023.
\newblock Treebanking user-generated content: a ud based overview of guidelines, corpora and unified recommendations.
\newblock \emph{Language Resources and Evaluation}, 57(2):493--544.

\bibitem[{Sterner and Teufel(2025{\natexlab{a}})}]{sterner2025code}
Igor Sterner and Simone Teufel. 2025{\natexlab{a}}.
\newblock \href {https://doi.org/10.18653/v1/2025.findings-acl.600} {Code-switching and syntax: A large-scale experiment}.
\newblock In \emph{Findings of the Association for Computational Linguistics: ACL 2025}, pages 11526--11533, Vienna, Austria. Association for Computational Linguistics.

\bibitem[{Sterner and Teufel(2025{\natexlab{b}})}]{sterner2025minimal}
Igor Sterner and Simone Teufel. 2025{\natexlab{b}}.
\newblock \href {https://doi.org/10.18653/v1/2025.acl-long.910} {Minimal pair-based evaluation of code-switching}.
\newblock In \emph{Proceedings of the 63rd Annual Meeting of the Association for Computational Linguistics (Volume 1: Long Papers)}, pages 18575--18598, Vienna, Austria. Association for Computational Linguistics.

\bibitem[{Tian et~al.(2024)Tian, Xia, and Song}]{tian2024large}
Yuanhe Tian, Fei Xia, and Yan Song. 2024.
\newblock Large language models are no longer shallow parsers.
\newblock In \emph{Proceedings of the 62nd Annual Meeting of the Association for Computational Linguistics (Volume 1: Long Papers)}, pages 7131--7142.

\bibitem[{Zhang et~al.(2025)Zhang, Hou, Gong, and Li}]{zhang2025self}
Ziyan Zhang, Yang Hou, Chen Gong, and Zhenghua Li. 2025.
\newblock \href {https://doi.org/10.18653/v1/2025.findings-emnlp.357} {Self-correction makes {LLM}s better parsers}.
\newblock In \emph{Findings of the Association for Computational Linguistics: EMNLP 2025}, pages 6749--6762, Suzhou, China. Association for Computational Linguistics.

\end{thebibliography}
